# Accelerated Training for Massive Classification via Dynamic Class Selection

Xingcheng Zhang,[1] Lei Yang,[1] Junjie Yan,[2] Dahua Lin[1]
[1]Department of Information Engineering, The Chinese University of Hong Kong
[2]SenseTime Group Limited
{zx016, yl016, dhlin}@ie.cuhk.edu.hk, yanjunjie@sensetime.com

## Abstract

Massive classification, a classification task defined over a vast number of classes (hundreds of thousands or even millions), has become an essential part of many real-world systems, such as face recognition. Existing methods, including the deep networks that achieved remarkable success in recent years, were mostly devised for problems with a moderate number of classes. They would meet with substantial difficulties, *e.g.* excessive memory demand and computational cost, when applied to massive problems. We present a new method to tackle this problem. This method can *efficiently* and *accurately* identify a small number of *"active classes"* for each mini-batch, based on a set of dynamic class hierarchies constructed on the fly. We also develop an adaptive allocation scheme thereon, which leads to a better tradeoff between performance and cost. On several large-scale benchmarks, our method significantly reduces the training cost and memory demand, while maintaining competitive performance.

## Introduction

Recent years have witnessed a wave of breakthroughs in AI areas (Shakirov 2016), thanks to the advances in deep learning and the explosive growth of datasets. Along with this trend, *massive classification* that involves an exceptionally large number of classes emerges as an important task. Such a task often arises in applications like face recognition (Sun, Wang, and Tang 2014) or language modeling (Chen, Grangier, and Auli 2015), when industry-level datasets are used.

Massive classification poses a number of new challenges, among which, the computational difficulties in training are perhaps the most prominent. Specifically, a contemporary classifier often adopts a deep network architecture, which typically comprises a series of transformation layers for feature extraction and a *softmax* layer that connects the top-level feature representation with per-class responses. The parameter size of this softmax layer is proportional to the number of classes. When training such a classifier, for each mini-batch of samples, the responses for *all* involved classes will be computed by taking the dot products between the class-specific weights and the extracted features. When there is a huge number of classes, the algorithm as described above will be faced with two significant difficulties: (1) The parameter size may go beyond the memory capacity, especially when the network is trained on GPUs with limited capacity. (2) The computational cost will be dramatically increased, even to a prohibitive level.

Existing efforts to tackle such difficulties (Mnih and Teh 2012; Jean et al. 2015; Wu et al. 2016) mostly originate from natural language processing, where one often needs to deal with large vocabularies. A common idea adopted by these methods is to focus on *frequent words*, based on *prior statistics* collected from the training corpus. However, it is often difficult to extend these methods along to other domains, as they require an essential condition to work, namely, the frequencies of different classes are highly imbalanced. This is not the case for applications like face recognition.

We conduct an empirical study when exploring this problem (see Sec. 3). In this study, we have two important observations: (1) For most cases, the samples of a class can only be confused with a small number of other classes. (2) When a softmax loss is used, for each sample, the signals back-propgated from this small subset of classes dominate the learning process. These observations suggest that for each mini-batch of samples, it suffices to perform the computation over a small fraction of classes, while excluding other classes whose impact is negligible. Motivated by this study, we explore a new approach to tackling the challenges of massive classification. Our basic idea is to develop a method that can identify a small number of *active classes* that can yield significant signals for each mini batch of samples. This method needs to be both *accurate* and *efficient*. To be more specific, it should be able to correctly identify those classes that are truly correlated with the given samples, but would not incur too much overhead. Satisfying both requirements at the same time is nontrivial.

Our efforts towards this goal consist of two stages. First, we derive an optimal class selector. Through experiments, we show that with *optimal selection* of active classes, the learned network can achieve the same level of performance with only 1% of classes selected at each iteration. However, the *optimal selection* requires computing responses for *all* classes, which in itself is overly expensive. Then in the second stage, we develop an *efficient* approximation of the *optimal selector* based on *dynamic class hierarchies*. The dynamic class hierarchies effectively capture the proximity among classes, with which one can approximately identify the *active classes* for a mini-batch with dramatically lower cost. Note that the class hierarchies are *dynamically* updated based on the weight vectors associated with individual classes. Hence, the selection of *active classes* takes into account *both* the sample features and the weight vectors. This distinguishes our method from the aforementioned methods that rely on prior statistics. Also, we observe that the average number of truly active classes tends to decrease as the training proceeds, and thus develop an *adaptive allocation scheme* accordingly, which leads to better tradeoff between cost and performance.

On large-scale benchmarks, LFW (Huang et al. 2007), IJB-A (Klare et al. 2015), and Megaface (Kemelmacher-Shlizerman et al. 2016), our method can remarkably reduce the training time and the memory demand.

## Related Work

Existing methods to tackle the challenges of massive classification primarily focus on reducing the cost of the softmax layer, as it constitutes the computational bottleneck. These methods roughly fall into three categories below.

**Approximated softmax**  An important idea explored previously is to devise approximations to the *softmax* function to reduce the computational cost. The *Hierarchical Softmax (HSM)* (Goodman 2001) is an early attempt along this line. HSM reformulates the multi-class classifier into a hierarchy of binary classifiers. During training, given a sample, it only has to traverse along a path from the root to the corresponding class, and thus the cost is reduced. This method has two limitations. First, it requires a prior distribution of the classes to build the hierarchy. More importantly, the resultant tree does not necessarily help discrimination, as the prior provides no information as to the proximity among classes.

**Frequency-based methods**  Another idea is to focus on the most frequent words, which is commonly seen in language modeling context. Schwenk proposed a continuous-space coding of words (2007), which prunes the output layer by just retaining a shortlist of the most frequent words. Le et al. proposed a structured output layer for neural network language models (2011), where the most frequent words are directly connected to the hidden layer while others are via binary trees. Jean et al. proposed to uses a predined proposal distribution to sub-sample classes in the softmax layer 2015. Chen, Grangier, and Auli proposed *Differentiated softmax (D-Softmax)* (2015), which partitions the words into different subsets based on frequencies. The words that occur more frequently will be associated with parameters of higher dimensions. Recently, Grave et al. proposed *Adaptive Softmax* (2016), which utilizes a short list to keep the most frequent words in the root of a two-layer tree. This method can notably reduce the training time with rare words excluded from a considerable part of the computation.

All the methods above exploit the *imbalanced distribution* of classes (*e.g.* the words in language models) to reduce computation. This, however, is not always the case in real-world applications. For example, in face recognition, all classes are equally important in general. For such problems, these methods are not suited.

**Noise contrastive learning**  For softmax computation, the major cost lies in computing the normalization constant, which has to sum over *all* classes. Gutmann and Hyvärinen proposed *Noise Contrastive Estimation (NCE)* (2010), an alternative way to estimate probabilistic distributions that circumvents the normalization issue. The basic idea is to replace the original maximum-likelihood objective by a binary logistic regression. *NCE* has been applied to cope with large vocabularies in language models (Mnih and Teh 2012; Mnih and Kavukcuoglu 2013; Vaswani et al. 2013). Word2Vec (Mnih and Teh 2012) also adopts this approach and obtains promising results. However, it has been repeatedly shown in previous work that promoting the contrast among classes is crucial for discriminative learning (Sun, Wang, and Tang 2014; Sun et al. 2014). Turning multi-class classification to binary logistic regression may result in weaker discriminative power and thus inferior performance.

**Key differences**  Whereas the proposed method adopts the idea of sub-sampling classes, it differs from previous work in several key aspects: (1) It does not rely on an imbalanced distribution of classes. Instead, it exploits the *proximity* among them and explicitly promotes the contrast among similar classes. This distinguishes it from both frequency-based methods and *NCE*-based methods. (2) The sampling is guided by *dynamic hierarchies*, which are adaptively updated on the fly during training, rather than being fixed in advance.

## Problem Statement

As a core topic in machine learning, classification has been extensively studied. In recent years, due to the remarkable success of deep learning (Schmidhuber 2015), deep networks have become very popular in classifier formulations. A classification pipeline typically consists of two stages, feature extraction and feature-based classification, as:

$$\mathbf{p} = \boldsymbol{\sigma}\left(\mathbf{W}\mathbf{x}\right), \quad \text{with } \mathbf{x} = F(\mathbf{o}; \boldsymbol{\theta}) \in \mathbb{R}^D. \quad (1)$$

Here, $\mathbf{o}$ denotes the observed input for a sample, $\mathbf{x} = F(\mathbf{o}; \boldsymbol{\theta})$ the extracted feature, and $\boldsymbol{\theta}$ the parameter of the feature extractor. Given a feature vector $\mathbf{x} \in \mathbb{R}^D$, the linear transform $\mathbf{W}\mathbf{x}$ turns it into *per-class responses*. Let $N$ be the number of classes, then $\mathbf{W}$ is a matrix of size $N \times D$, where each row $\mathbf{w}_i$ is the weight vector for the $i$-th class.

Following the practice in the seminal work (Krizhevsky, Sutskever, and Hinton 2012), classification models usually uses a softmax function $\boldsymbol{\sigma}$ to transform $\mathbf{W}\mathbf{x}$ into $\mathbf{p} \in \mathbb{R}^N$, a vector of posterior probabilities. The learning objective is often defined to be maximizing the joint probability of the ground-truth classes. Specifically, the *softmax function* is defined as $\boldsymbol{\sigma}(\mathbf{y}) = [\sigma_1(\mathbf{y}), \ldots, \sigma_N(\mathbf{y})]$, where

$$\sigma_i(\mathbf{y}) = e^{y_i} / \sum_{j=1}^{N} e^{y_j}. \quad (2)$$

The major cost in this computation lies in the linear transform $\mathbf{Wx}$. Both the time complexity and the space complexity increases linearly as the number of classes $N$ increases.

This may not be a significant issue for problems of moderate scale. However, as machine learning techniques are increasingly used in the context with large-scale data, the problem of massive classification emerges and even becomes a critical issue in certain applications. An important example is face recognition. As shown in (Sun, Wang, and Tang 2014), training on a dataset with a massive number of classes is crucial for obtaining strong performance. Large datasets released recently even comprise nearly a million classes (Kemelmacher-Shlizerman et al. 2016).

Such datasets post significant computational challenges. On one hand, training on datasets as such requires huge computing power. Therefore, GPU acceleration is needed. On the other hand, current GPUs only come with limited memory capacity, *e.g.* the memory capacity of *Tesla P100* is up to 16 GB. When training a network on multiple GPUs, this issue is even more severe – synchronizing the massive weight matrix $\mathbf{W}$ would incur significant communication overhead, thus adversely affecting the overall efficiency.

### An Empirical Study of Softmax

To explore an effective way to reduce the cost of the softmax layer for massive classification, we conduct an empirical study. In particular, we hypothesize that for each given sample, it may yield positive responses for only a small number of classes, even when the total number is very large. To verify this, we introduce a quantity called *top-K cumulative probability*, which is defined as the sum of predicted probabilities, *i.e.* the probability values produced by softmax, for top $K$ classes:

$$CP_K(\mathbf{x}) \triangleq \sum_{i \in \mathcal{T}_K(\mathbf{y})} \sigma_i(\mathbf{y}), \quad \text{with } \mathbf{y} = \mathbf{Wx}, \quad (3)$$

where $\mathcal{T}_K(\mathbf{y})$ is the set of those $K$ classes with highest responses. A high value of $CP_K$ indicates that the predicted probabilities concentrate on top $K$ classes. Fig 1(a) shows the change of average $CP_K$ along with the training epochs, obtained on *MS-Celeb-1M*, a large dataset for face recognition that contains about $87K$ classes (Guo et al. 2016). We can see that as the training proceeds, the probabilities gradually become more and more concentrated. Particularly, after 23 epochs, over $95\%$ of the probability mass fall onto the top-1000 classes, which is only $1.15\%$ out of all.

On the other hand, the softmax loss provides feedback signals via gradients. For a training sample from class $c$, the gradient at $\mathbf{w}_i$, the $i$-th row of $\mathbf{W}$, is given by

$$\mathbf{g}_i := \frac{\partial(-\log \sigma_c(\mathbf{Wx}))}{\partial \mathbf{w}_i} = \begin{cases} \sigma_i(\mathbf{y})\mathbf{x} & i \neq c, \\ (\sigma_i(\mathbf{y}) - 1)\mathbf{x} & \text{otherwise.} \end{cases} \quad (4)$$

We can see that it is proportional to $\sigma_i(\mathbf{y})$ (for $i \neq c$), the predicted probability for $i$. Hence, if the predicted probabilities concentrate on the top $K$ classes, then the gradients should also concentrate on those classes.

We verify this empirically. Specifically, we introduce another quantity called *normalized top-K cumulative gradient energy*, which is defined as:

$$NCG_K(\mathbf{x}) = (\frac{\mathbf{g}^T \hat{\mathbf{g}}}{||\mathbf{g}|| \cdot ||\hat{\mathbf{g}}||})^2 \quad \text{with } \hat{g}_i = g_i \mathbb{1}_{i \in \mathcal{T}_K(\mathbf{y})}, \quad (5)$$

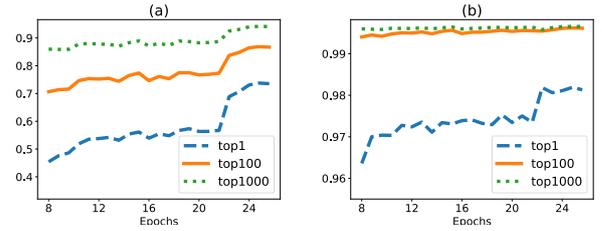

Figure 1: The curves of (a) average $CP_K$ values and (b) average $NCG_K$ values vs. the number of elapsed training epochs. The values are obtained on *MS-Celeb-1M*.

It measures how much the gradients are concentrated on the top $K$ classes. Fig. 1(b) shows the change of average $NCG_K$ along with the training epochs. It demonstrates that the gradients are highly concentrated on a small fraction of top classes. These observations suggest that we can still effectively learn the classifier if we restrict the softmax loss to those top classes that dominate the probabilities and gradients. In what follows, we refer to them as the *active classes*.

## Methodology

The basic idea of our method is to select a small number of *active classes* for each mini-batch of samples and compute a *selective softmax* and the gradients based on them. In this section, we first provide an overview of the entire pipeline, and then proceed with a detailed discussion of *how to select active classes*. Particularly, we begin with an optimal but costly scheme, and then develop an efficient approximation based on *dynamic class hierarchies*. Futhermore, we introduce an *adaptive allocation scheme*, which adjusts the algorithmic parameters as the network status changes, so as to strike a better balance between performance and cost. Finally, we will present useful details in our implementation.

### Selective Softmax

The empirical study in previous section shows that a small number of *active classes* dominate not only the predicted probabilities but also the gradients for backpropagation. Inspired by this, we propose a variant of softmax, called *selective softmax*, to approximate the *full softmax*. The *selective softmax* is defined as follows:

$$\mathcal{E}_{\mathcal{S}(\mathbf{x})}\left[\boldsymbol{\sigma}_{\mathcal{S}(\mathbf{x})}(\widetilde{\mathbf{W}}\mathbf{x})\right], \text{ with } \widetilde{\mathbf{W}} = \mathcal{F}_{\mathcal{S}(\mathbf{x})}(\mathbf{W}). \quad (6)$$

Here, $\mathcal{S}(\mathbf{x})$ denotes the set of *active classes* selected for the given sample $\mathbf{x}$. Given $\mathcal{S}(\mathbf{x})$, the computation proceeds in several steps, as follows: (1) Derive a sub-matrix $\widetilde{\mathbf{W}}$ from the current weight matrix $\mathbf{W}$, which contains the rows corresponding to the classes in $\mathcal{S}(\mathbf{x})$. Let $M$ be the number of selected classes, then $\widetilde{\mathbf{W}} \in \mathbb{R}^{M \times D}$. The notation $\mathcal{F}_{\mathcal{S}(\mathbf{x})}$ indicates this operation. (2) Compute $\widetilde{\mathbf{W}}\mathbf{x}$, the responses for the selected classes. (3) Turn $\widetilde{\mathbf{W}}\mathbf{x}$ into an $M$-dimensional probability vector via $\boldsymbol{\sigma}_{\mathcal{S}(\mathbf{x})}$, a softmax function restricted to $\mathcal{S}(\mathbf{x})$. (4) Expand this probability vector into an $N$-dimensional vector, setting the entries corresponding to the unselected classes to zeros. The notation $\mathcal{E}_{\mathcal{S}(\mathbf{x})}$ indicates this operation. Note that the expanded vector is a valid probability vector over all $N$ classes.

The *selective softmax* above is equivalent to forcing the probabilities to be zeros for *non-active classes*, while renormalizing those of the active ones. If the active classes are correctly selected, this would be a very close approximation to the full softmax, as the original probabilities of *non-active classes* are indeed nearly zeros. Consequently, the resultant gradients also closely approximate the original gradients.

Using *selective softmax*, both the computational complexity and the memory demand are reduced from $O(ND)$ to $O(MD)$. When $M \ll N$, this would be a dramatic reduction. Our experiments show that with a proper selection scheme, we can maintain the same level of performance with only 1% of the classes selected at each iteration.

**Optimal Selection of Active Classes**

*Selective softmax* provides a good approximation to the full softmax with substantially reduced cost. However, the effectiveness of this approach hinges on whether we can *accurately* and *efficiently* selection the *active classes*. This is the key challenge of our work.

For a softmax function, a class with higher response value $\mathbf{w}_i^T \mathbf{x}$ also has higher predicted probability $\sigma_i(\mathbf{y})$ and greater gradient norm $\|\mathbf{g}_i\|$. Therefore, with a fix quota $Q$, the optimal strategy to select *active classes* for a given sample $\mathbf{x}$ is to choose those classes that yield the highest responses.

As we will show empirically in next section, this simple method can *accurately* find the active classes. Using this scheme, a very small quota $Q$ is usually enough for training a classifier with competitive performance. This method, however, has a serious drawback – it is too expensive. To find the $M$ classes with highest responses, one has to compute $\mathbf{W}\mathbf{x}$ entirely in order to derive the responses for *all* classes. The complexity of this procedure is $O(ND)$, at the same level as the original full softmax function. Hence, this selection scheme is not a viable choice in practice, as its significant complexity defeats the very purpose of our work, that is, to reduce the computational cost.

Nonetheless, the optimal scheme above reveals two key factors that affect the selection: the weight matrix $\mathbf{W}$ and the sample feature $\mathbf{x}$. For $\mathbf{W}$, its columns can be viewed as the embeddings of classes in a $D$-dimensional space, which capture the proximity among classes. The proximity is crucial for identifying classes that can be easily confused. The sample feature $\mathbf{x}$, on the other hand, provides instance-level information. Due to large intra-class variation, the subset of active classes may significantly vary across instances, even within the same class. Also, as the extracted feature $\mathbf{x}$ for each instance changes over iterations, it is important to use the *updated features* for active classes selection.

**Approximated Selection**

---

**Algorithm 1** Build Hashing Tree
**Input:** $\mathbf{W}$, root
**Output:** Tree
1: **while** $|\mathbf{W}| > B$ **do**
2:     Randomly sample $\mathbf{w}_i, \mathbf{w}_j \in \mathbf{W}, i \neq j$
3:     Get the normal vector $\mathbf{h} = \frac{\mathbf{w}_i + \mathbf{w}_j}{2}$
4:     $\mathbf{W}_l = \arg_{\mathbf{w}_i \in \mathbf{W}}(\mathbf{w}_i^T \mathbf{h} \geq 0)$
5:     $\mathbf{W}_r = \arg_{\mathbf{w}_i \in \mathbf{W}}(\mathbf{w}_i^T \mathbf{h} < 0)$
6:     Build Hashing Tree ($\mathbf{W}_l$, root→left)
7:     Build Hashing Tree ($\mathbf{W}_r$, root→right)
8: **end while**
9: **return** Tree

---

We propose to use a *Hashing Forest (HF)* to approximate the optimal selection scheme. The gist of this method is to partition the space of weight vectors into small cells via recursive partitioning. Specifically, it begins with the entire space as a cell, and recursively applies random partitioning as follows. At each iteration, it chooses a cell with more than $B$ points, randomly picks a pair of points therein, and then computes a hyperplane that separates the chosen points with maximum margin. This hyperplane will split this cell into two. The procedure continues until no cells contain more than $B$ points. Note that this recursive partitioning procedure essentially builds a binary tree of the cells. Overall, the tree building procedure is listed in Algorithm 1.

As each cell is partitioned in a stochastic way, a single hashing tree might not be enough to capture the proximity among classes reliably. To improve the accuracy of search, we build $L$ hashing trees, which together form a *hashing forest*. The number of trees $L$ can be determined empirically based on the trade-off between cost and accuracy.

Given a query feature $\mathbf{x}$ and a required quota $Q$, we walk along the tree starting from the root node. At each iteration, it will choose a branch according to which side of the boundary that it falls in. This stops when it hits a node (not necessarily a leaf cell) that contains less than $Q$ points. Then, it selects all the classes under its parent (whose size is at least $Q$). This procedure will be repeated for each hashing tree in the forest. Among all resultant classes pooled from individual trees, we choose $Q$ that are closest to $\mathbf{x}$ (by cosine distance) to form $\mathcal{S}(\mathbf{x})$.

On average, the depth of a hashing tree is $\log \frac{N}{B}$. At each level (except in the leaf cell), all points have to be dispatched to either side of the boundary by geometric calculation. Hence, the overall complexity for building a hashing tree is $O(N \log \frac{N}{B})$. As $B \ll N$ in practice, we may consider the complexity as $O(N \log N)$. For query, the complexity of top-down traverse is $O(\log \frac{N}{Q})$ on average. When it hits the target node, it has to compute the cosine distances from $\mathbf{x}$ to all points therein, for which the complexity is $O(Q)$. Hence, the overall complexity for a query (along a tree) is $O(\log \frac{N}{Q} + Q)$. For a forest, the complexity for both building and querying grows linearly with $L$. However, the time can be reduced by parallelizing the operation on different trees.

## Adaptive Allocation

The empirical study in Section 3 shows that both the predicted probabilities and the gradients become more concentrated as the training proceeds, due to the increasing discriminative power of the model. Inspired by this, we propose an improved training scheme, which adaptively adjusts the parameters that affect the use of computing resources (*e.g.* allowing more active classes in early epochs), so as to pursue a better balance between performance and cost.

This scheme controls three parameters: (1) $M$, the number of active classes for each iteration, (2) $L$, the number of hashing trees, and (3) $T$, the interval of structure update. In particular, we will rebuild the hashing forest every $T$ iterations in order to stay updated. Generally, increasing $M$ and $L$ while reducing $T$ can improve the performance, but at the expense of higher computing cost. Our strategy adjusts $M$, $L$, and $T$ periodically, based on several heuristics. Specifically, we divide the entire training process into a few phases, each spanning a number of epochs. At the beginning of a phase, we reset these design parameters according to the updated states of the network, as follows.

First, we set $M$ to be the minimum number such that the average top-$M$ cumulative probability is above a threshold $\tau_{cp}$, where $\tau_{cp}$ increases linearly with the number of iterations. Second, we increase the update interval $T$ linearly, as the training process stabilizes over time. Third, we initialize the number of trees $L$ to be a small number, and increase $L$ linearly over time. The rationale behind is that the selection need not be very accurate in early stages as everything is volatile, and more accurate selection is needed as the probabilities become more concentrated in later stage.

## Implementation Details

For the sake of clarity, the algorithm discussed above only involves one instance. In practice, we use a mini-batch of instances for each iteration. Let $\mathcal{X}$ denotes a set of extracted feature $\mathbf{x}$ for the current mini-batch, then $\mathcal{S}(\mathcal{X})$ is a union of $\mathcal{S}(\mathbf{x})$. Also, it is often convenient to have a maximum size for the set of active classes used in each iteration. When the cardinality of $\mathcal{S}(\mathcal{X})$ is greater than the maximum size, sampling based on the renormalized probabilities of active classes will be applied.

Besides, in our approach, since only a small number of rows in $\mathbf{W}$ will be used and updated in each iteration, we adopt a client-server architecture (Xie et al. 2016) to support efficient *sparse update*. Particularly, it maintains the entire $\mathbf{W}$ in a large-capacity memory (*e.g.* CPU Ram) on the server. When a subset of active classes is selected, it will retrieve the corresponding sub-matrix $\widetilde{\mathbf{W}}$ and cache it in the client's GPU. When the updates are done, the gradients will be sent back to the server. Our implementation adopts the parameter server design (Ho et al. 2013), which allows sparse updates to be done asynchronously.

# Experiments

We test our method on three benchmarks on face recognition/verification, which is the application that motivates this work. We not only compare it with various methods, but also investigate how different factors influence the performance and cost, via a series of ablation studies.

## Experimental Settings

**Datasets** Following the convention in face recognition, we train networks on large training sets that are *completely disjoint* from the testing sets, namely the identities (*i.e.* classes) used for testing are excluded from the training set.

Specifically, two large datasets below are used for training: **(1) MS-Celeb-1M** (Guo et al. 2016). This dataset consists of $100K$ identities, each with about 100 facial images on average. In total, the dataset contains $10M$ images. As the original identity labels are extracted *automatically* from webpages and thus are very noisy. We clean up the annotations manually, resulting in a reliable subset that contains $4.6M$ images from $87K$ classes. **(2) Megaface (MF2)** (Kemelmacher-Shlizerman et al. 2016). This is one of the largest annotated training sets for face recognition that are publicly available, which contains $4.7M$ images from $672K$ identities. Note that these two datasets differ significantly in the number of identifies ($87K$ *vs.* $672K$) and the average number of images per identity ($100$ *vs.* $7$). Comparing the results on both datasets reveals the differences among various methods under different settings.

The trained networks are then tested on three testing sets: **(1) LFW** (Huang et al. 2007), the *de facto* standard testing set for face verification under unconstrained conditions, which contains $13,233$ face images from $5,749$ identities. **(2) IJB-A** (Klare et al. 2015), which contains $5,712$ face images from $500$ identities. **(3) Megaface & Facescrub**, the largest public benchmark for face recognition, which combines the gallery sets from both Megaface (Kemelmacher-Shlizerman et al. 2016) (with $1M$ images from $690K$ identities), and Facescrub (Ng and Winkler 2014) (with $100K$ images from $530$ identities).

**Metrics** We assess the performance on two tasks, namely *face identification* and *face verification*. Face identification is to select top $k$ images from the gallery, where the performance is measured by the *top-k hit rate*, *i.e.* the fraction of predictions where the true identity occurs in the top-$k$ list. Face verification is to determine whether two given face images are from the same identity. We use a metric widely adopted in practice, namely the *true positive rate* under the condition that the *false positive rate* is fixed to be $0.001$.

**Networks** We conduct two series of experiments, with different network architectures. First, we experiment over *MS-Celeb-1M* training set with various methods under different settings, based on Hynet, a variant of VGG (Simonyan and Zisserman 2014) with certain parts optimized for higher efficiency. Using a network of moderate size like this allows us to conduct extensive studies within a reasonable budget. To further study how different methods work with very deep networks, we conduct another series of experiments for selected methods using ResNet-101 (He et al. 2016), over a larger training dataset, namely the union of *MS-Celeb-1M* and *Megaface (MF2)*. For all settings, the networks are trained using SGD with momentum. The mini-batch sizes are set to 512 and 256 respectively for Hynet and ResNet-101.

| Method | IJB-A | *Megaface (FaceScrub+1M)* |
|---|---|---|
| Full softmax | 0.860 | 0.630 |
| Random (1K) | 0.771 | 0.515 |
| PCA (1K) | 0.773 | 0.524 |
| Kmeans (1K) | 0.765 | 0.491 |
| Optimal (1K) | 0.863 | 0.625 |
| HF (1K) | 0.856 | 0.599 |
| NCE | 0.313 | - |
| HSM | 0.488 | - |

Table 1: Quantities in IJB-A column are *TPR* when *FPR* is 0.001, and in Megaface identification task is the top-1 accuracy.

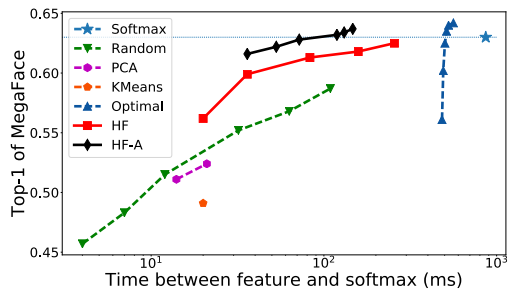

Figure 2: Face identification accuracy vs. computing cost for different methods. The points towards the top-left corner indicate high performance with low cost. Note that the x-axis is in *log-scale*.

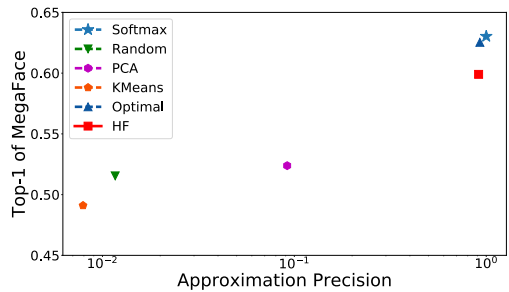

Figure 3: Face identification accuracy vs. approximation precision for different methods. Note that the x-axis is in *log-scale*.

## Method Comparison

We compare the proposed method with a series of baselines. These methods are briefly described below.
**(1) Softmax**, the full version of softmax, which involves *all* classes as active classes at each iteration.
**(2) Random**, a naive method, which randomly picks a fixed number of classes for softmax computation at each iteration.
**(3) PCA**, a simple hashing method. It projects the weight vectors to a low-dimensional space and encodes it into a binary hash code based on the polarity of each entry of the projected vector. For a sample from class $c$, the classes which share the same hash code will be selected.
**(4) KMeans**, another simple way to find active classes. It clusters all classes into $1024$ groups based on the corresponding weight vectors. For a sample from class $c$, the group that contains $c$ will be chosen.
**(5) Optimal**, the optimal selection scheme, which computes $\mathbf{Wx}$ entirely and chooses those classes that yield the highest responses as the active classes. This provides a performance upper bound, but it is very expensive.
**(6) HF**, the basic version of the proposed method, which approximates the optimal selection scheme using a hashing forest that will be updated periodically.
**(7) HF-A**, an advanced version of the proposed method that uses the *adaptive allocation scheme* to adjust the use of computing resources, in order to strike a better balance.

**Results** We train networks using different methods on *MS-Celeb-1M* and test them on *Megaface & Facescrub*. Tab. 1 shows some quantity results for different methods. And Fig. 2 compares their *performance* (against *cost*). Here, the *performance* is measured by top-1 identification accuracy, while the *cost* includes the time for both class selection and computing the *restricted* softmax and the gradients.

The results show: (1) For *random sampling*, the performance increases slowly as the number of active classes grows. This suggests that *random sampling* is not a very efficient way for selecting active classes. (2) *Cluster-based* methods like *K-means* and *PCA* do not perform very well, which suggests that only using the clusters of weight vectors while neglecting the sample feature $\mathbf{x}$ is not enough to obtain good performance. (3) The *optimal selection* delivers very high performance, even slightly surpassing *full softmax*. But this is at the expense of severe overhead in class selection, as it requires computing responses for *all* classes. (4) Our method *(HF)* clearly outperforms others by a large margin. At the same cost, it can yield considerably higher performance. On the other hand, to reach the same performance, it requires substantially lower cost (sometimes by even nearly an order of magnitude). (5) *HF-A* achieves clear improvement over the basic version, which clearly shows the merit of the *adaptive allocation* strategy. It is noteworthy that it can surpass the performance of *full softmax* with only $10\%$ of the cost. This is a remarkable improvement. (6) The results of *NCE* and *HSM* are far inferior to a typical method. Such results may suggest that these two methods are not quite suitable for massive classification problems.

Fig. 3 shows the performance against the approximation precision, *i.e.* the average overlap between the selected classes and the optimal selection. We see that better approximation generally comes with higher performance. The classes selected by the proposed *HF* are very close to the optimal subset (much more precise than other methods), and consequently obtain notably higher performance.

## Large Scale Experiment

We also conduct a large-scale experiment with ResNet-101, using the union of *MS-Celeb-1M* and *Megaface (MF2)* as the training set, which contains $750K$ classes in total. The training is done on a server with 8 NVIDIA Titan X GPUs. In this experiment, *HF-A* can reduce the training time per it-eration from 3.5 s to 1.5 s, which is a speedup by 60%. Note that this is the overall speedup of for the entire network (not just the softmax layer). Also the overall memory consumption on each GPU is reduced from $10.8G$ to $8.2G$. This is due to the dramatic reduction of the GPU memory demand from the softmax layer, as shown in Table 2. It also shows that our method can achieve a performance comparable to full softmax, with only 1% of the classes selected for each iteration. This result suggests that our method can scale well to massive classification problems.

| Method | Cost (s) | Memory (Mb) | Accuracy (%) |
|---|---|---|---|
| Full Softmax | 3.5 | 2511 | 64.7 |
| Random | 1.45 | 9 | 50.2 |
| HF-A | 1.5 | 9 | 63.9 |

Table 2: Performance vs cost in the large-scale experiment.

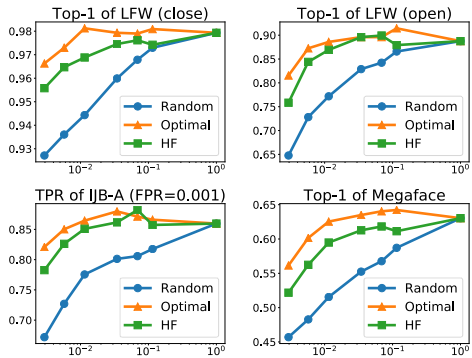

Figure 4: Performance vs. the number of active classes $M$ for Random, Optimal, and HF, on different datasets.

**Ablation studies**

**Number of Active Classes** $M$  The number of active classes is an important factor that influences the final performance. Generally, larger number active classes can result in better performance. This is confirmed by Fig. 4. Here, we compare three methods, *Random*, *Optimal Selection*, and *HF*. The former two respectively provide a lower and upper bounds in terms of performance. For all these methods, we can see performance increases as the number of *active classes* per iteration increases. But for *Optimal Selection*, the increase is much faster than *Random*, which indicates the importance of accurately identifying the *active classes*. The performance of proposed method *HF* also comes close to *Optimal Selection* in terms of performance given $M$, but it achieves close performance with substantially lower cost.

**Number of Hashing Trees** $L$  The number of hashing trees directly relates to the accuracy of approximating the *Optimal Selection*. Fig. 5 shows that the performance increases with the number of trees. There is a large performance gap between $L = 5$ and $L = 50$, but the performance gradually saturates as $L$ increases beyond 100.

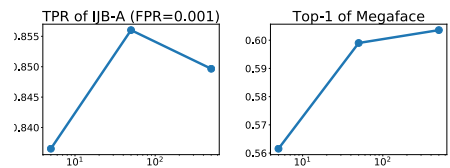

Figure 5: Performance vs. number of hashing trees $L$.

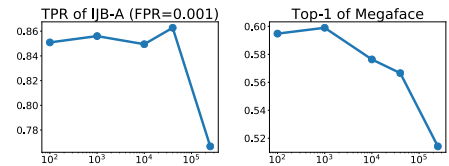

Figure 6: Performance vs. interval of structure update $T$

**Interval of Structure Update** $T$  The cost on building the hashing forest can be hidden by updating the structure in the background. Hence, this cost is generally not an issue. However, it is still interesting to examine the influence of the interval $T$ on performance. Fig.6 shows the performance do not change much when $T$ ranges from 100 to 1000 iterations, but becomes more sensitive to $T$ in the range between 1000 and 10000. The rightmost point is an extreme case that only builds the forest at the beginning of the training, the result of which is close to the result of random sampling.

**Threshold of Adaptive Allocation** $\tau_{cp}$  The performance is not very sensitive to the threshold. In our experiments, when $\tau_{cp}$ changes from 0.7 to 0.9, the accuracy improves by about 2%. Generally, higher threshold results in better performance, but the reward diminishes when the threshold increases beyond 0.9.

## Conclusion

This paper presents a new method to tackle the computation difficulty of the softmax layer for massive classification problems. Particularly, we develop an efficient method based on dynamic class hierarchies to approximate the optimal selection, which can *accurately* and *efficiently* identify the active classes for each mini-batch of samples. We also develop an adaptive allocation scheme on top that achieves better tradeoff between performance and cost by adaptively adjusting the allocation of computing resources. Experiments on large benchmarks show that the proposed method can speed up the overall training procedure by 60% and reduce the GPU memory demand by 24% without compromising performance. Note that this is achieved by only optimizing the softmax layer (without modifying other layers).

## Acknowledgement

This work is partially supported by the Big Data Collaboration Research grant from SenseTime Group (CUHK Agreement No. TS1610626), the General Research Fund (GRF) of Hong Kong (No. 14236516).